\documentclass[]{article}
\usepackage[whole]{bxcjkjatype}

\usepackage{soul}
\usepackage[a4paper]{geometry}
\usepackage{mtsummit2023}
\usepackage{times}
\usepackage{latexsym}
\usepackage{url}
\usepackage{natbib}
\usepackage{layout}
\usepackage[hang,marginal]{footmisc}
\usepackage{graphicx}

\usepackage{mathptmx}
\usepackage{xcolor}
\usepackage{color}
\usepackage{float}
\usepackage{subcaption,booktabs}
\usepackage{listings}
\usepackage{times} 
\usepackage{latexsym} 
\usepackage{caption}
\usepackage{amssymb,amsmath,amsthm,enumitem}

\usepackage[utf8]{inputenc}
\usepackage[T1]{fontenc}
\usepackage{microtype}
\usepackage{multirow}
\usepackage{amsmath}
\usepackage{booktabs,caption}
\usepackage[flushleft]{threeparttable}
\usepackage{booktabs}
\usepackage{tablefootnote}
\usepackage[defaultcolor=red!70!cyan]{changes}

\newcommand\hlj{\bgroup\markoverwith
  {\textcolor{yellow}{\rule[-.5ex]{2pt}{2.5ex}}}\ULon}

\begin{document}

\title{\bf{Context-aware Neural Machine Translation
for English-Japanese Business Scene Dialogues}}
 \author{\name{\bf{Sumire Honda} }  \hfill  \addr{pu.sumirehonda@gmail.com}\\
 \addr{Computational Linguistics, University of Potsdam,
         Potsdam, Germany}\\
         \name{\bf Patrick Fernandes} \hfill \addr{pfernand@cs.cmu.edu} \\
         \addr{Language Technologies Institute, Carnegie Mellon University, Pittsburgh, USA}\\
         \addr{Instituto de Telecomunicações, Instituto Superior Técnico, Lisbon, Portugal}\\
         \name{\bf{Chrysoula Zerva} } \hfill \addr{chrysoula.zerva@tecnico.ulisboa.pt}  \\
         \addr{Instituto de Telecomunicações, Instituto Superior Técnico, Lisbon, Portugal}\\
}

\maketitle
\pagestyle{empty}

\begin{abstract}
Despite the remarkable advancements in machine translation, the current sentence-level paradigm faces challenges when dealing with highly-contextual languages like Japanese. In this paper, we explore how \textit{context-awareness} can improve the performance of the current Neural Machine Translation (NMT) models for \textit{English-Japanese business dialogues} translation,
and what kind of context provides meaningful information to improve translation.
As business dialogue involves complex discourse phenomena but offers scarce training resources, we adapted a pretrained mBART model, finetuning on multi-sentence dialogue data, which allows us to experiment with different contexts. We investigate the impact of larger context sizes and propose novel context tokens encoding extra-sentential information, such as speaker turn and scene type.
We make use of \textit{Conditional Cross-Mutual Information} (CXMI) 
to explore how much of the context the model uses and generalise CXMI to study the impact of the \textit{extra-sentential context}.
Overall, we find that models leverage both preceding sentences and extra-sentential context (with CXMI increasing with context size) and we provide a more focused analysis on honorifics translation. Regarding translation quality, increased source-side context paired with scene and speaker information improves the model performance compared to previous work and our context-agnostic baselines, measured in BLEU and COMET metrics.
~\footnote{Code available at: \url{https://github.com/su0315/discourse_context_mt}}
\end{abstract}

\section{Introduction}
Traditionally NMT models such as Transformers \citep{10.1145/3441691} approach the task of machine translation (MT) focusing on individual sentences without considering the surrounding information, such as previous utterances or underlying topics.
As a result, the output often lacks discourse coherence and cohesion, which is problematic for MT applications such as chat translation systems \citep{farajian-etal-2020-findings,bawden-etal-2018-evaluating}. 
Thus, it is still an open research question to what degree these models can take advantage of contextual information to produce more accurate translations.  

To answer this question, several context-aware NMT \citep{ tiedemann-scherrer-2017-neural,voita2019good,maruf2019selective,xu2021efficient} studies have been conducted by adding surrounding sentences to the models and testing if it helps to capture better specific linguistic phenomena requiring context (e.g. coreference resolution). 
However, 
{there is limited work} on discourse or dialogue datasets, and most of {it is focused }
on high-resource or Indo-European (IE) languages \citep{2021}. 
Therefore, there is a need to investigate how well do the proposed approaches capture discourse phenomena in non-IE or low-resource languages. 

This work aims to address the aforementioned gap by focusing on English-Japanese (En-Ja) translation for business dialogue scenarios in order to examine if current context-aware NMT models \citep{tiedemann-scherrer-2017-neural} actually use the additional context, and what kind of context is useful regarding the translation of linguistic phenomena pertaining to Japanese discourse,
such as honorifics.  We specifically propose the use of novel extra-sentential information as additional context and show that it improves translation quality. 
Overall, the main contributions of this study are threefold: (1) We demonstrate that it is possible to adapt a (non-context-aware) large pretrained model (mBART; \cite{liu-etal-2020-multilingual-denoising, tang2021multilingual}) to attend to context for business dialogue translation and propose an \textbf{improved attention mechanism} (CoAttMask) with significant performance gains for source-side context, even on small datasets; (2) we propose \textbf{novel extra-sentential information} elements such as speaker turn and scene type, to be used as additional source-side \textbf{context}; and
 (3) we compare the use of context between our context-aware models using CXMI \citep{fernandes-etal-2021-measuring}, a mutual-information-based metric and perform a more focused analysis on the translation of \textbf{honorifics}.

\section{Related Work}
\subsection{Context-aware MT}
\label{sec:context-aware}
Context-aware MT lies between sentence-level MT and document-level MT, as the former assumes the translation of a single sentence from source to target language with no other accessible content, and the latter implies the translation of a sequence of sentences from a document, assuming access to the whole document. 
Context-aware MT lies close to the definition of document-level MT, as it requires access to context either in the form of preceding sentences or other type of information regarding the topic and setup of the text to be translated, that can aid in its translation.

Several methods using a transformer-based architecture \citep{NIPS2017_3f5ee243} have been proposed for context-aware NMT, frequently categorised into single-encoder and multi-encoder models \citep{sugiyama-yoshinaga-2019-data}.  
Single-encoder models concatenate the source sentence with (a) preceding sentence(s) as the contexts, with a special symbol to distinguish the context and the source or target in an encoder \citep{tiedemann-scherrer-2017-neural}.
Multi-encoder models pass the preceding sentence(s) used as context through a separate encoder modifying the Transformer architecture \citep{voita-etal-2018-context, tu-etal-2018-learning}. 
According to \citet{sugiyama-yoshinaga-2019-data}, the observed performance gap between the two models is marginal, but the single-encoder models are relatively simpler architectures without modifying sequence-to-sequence transformers. 

Apart from concatenating preceding sentences on the source-side, some works focus on the target-side context, i.e., show some benefits from attempting to decode multiple sequential sentences together \citep{Su2019Exploiting,mino-etal-2020-effective}. Depending on the use-case, source-side, target-side, or a combination of contexts has proven beneficial \citep{agrawal2018contextual,chen2021improving,fernandes-etal-2021-measuring}. Additionally, some works focused more on context related to discourse phenomena, with \cite{liang2021modeling} proposing the use of variational autoencoders to model dialogue phenomena such as speaker role as latent variables \citep{liang2021towards}. We examine here a simpler approach, that directly encodes such speaker and scene information and allows the model to use it as additional context. In more recent work, the impact of pretraining on larger out-of-domain (OOD) data has also been studied to aid in downstream MT tasks with limited resources \citep{voita2019good,liang2022scheduled}.

For English-Japanese translation, there have been some context-aware NMT studies that used variations of single-encoder models in the news 
and dialogue domain \citep{sugiyama-yoshinaga-2019-data, ri-etal-2021-zero, rikters-EtAl:2020:WMT}. Specifically for dialogue, \citet{rikters-EtAl:2020:WMT} experimented with context-aware MT that employs source-side factors on Ja-En (Japanese-English) and En-Ja (English-Japanese) discourse datasets. 
They propose to concatenate the preceding sentence(s) from the same document followed by a tag-token to separate the context from the original sentence and use binary token-level factors on top of this to signify whether a token belongs to the context or source sentence.

\subsection{Japanese Honorifics in NMT}

For into-Japanese MT, specific discourse phenomena such as honorifics constitute a core challenge when translating from languages that do not include such phenomena, like English \citep{hwang2021context, sennrich2016controlling}.
Japanese honorifics differ to English because different levels of honorific speech are used to convey respect, deference, humility, formality, and social distance, using different types of verbal inflexions. Besides, the desired formality is decided depending on social status and context and may involve more extensive changes in utterances compared to other languages \citep{fukada2004universal}.  
\citet{feely-etal-2019-controlling} proposed formality-aware NMT, conditioning the model on a manually selected formality level to evaluate honorifics. 
They evaluate the formality level of the translated sentences using their formality classifier, showing improvements.
Instead of explicitly selecting the formality level, we  evaluate the impact of our context representations on the correct translation of honorifics, inspired by \cite{yin2021does}.\looseness=-1

\section{Datasets}
\label{sec:data}
We use Business Scene Dialogue corpus (BSD) \citep{rikters-etal-2019-designing} as the main dataset.
Additionally, only to compare the performance in a certain setup with the main dataset, we also use AMI Meeting Parallel Corpus (AMI) \citep{rikters-EtAl:2020:WMT} as a supplemental dataset.
They are both document-level parallel corpora consisting of different scenes (dialogue sequence scenarios) or meetings and include both out-of-English and into-English translations, of which we use the English-Japanese translation direction. We focus our analysis on the BSD dataset, as it contains more scenarios and extra-sentential information which we use as additional context. 

In {the main dataset} BSD, each document consists of a business scene with a scene tag (face-to-face, phone call, general chatting, meeting, training, and presentation), and each sentence has speaker information that indicates who is speaking.
Contents of BSD are originally written either in English or Japanese by bilingual scenario writers who are familiar with business scene conversations and then translated into the other language to {create} a parallel corpus.

As for AMI, the contents are translations to Japanese from 100 hours of meeting recordings in English. {Since it originates from naturally occurring dialogue} it contains shorter utterances than BSD, {including multiple single-word sentences with filler and interjection words.} The data split statistics for BSD and AMI are shown in Table~\ref{tab:split}. 
The domain of BSD and AMI is similar, however, AMI does not include scene information and the number of documents (scenarios) is smaller.

\begin{table}[ht!]
\centering
\begin{tabular}{ccccccccc}
\toprule 
& \textbf{BSD} & Train & Dev & Test & \textbf{AMI} & Train & Dev & Test \\ 
\cmidrule(l{2pt}r{2pt}){3-5} \cmidrule(l{2pt}r{2pt}){7-9}  
Sentences & & 20,000   & 2051        & 2120  & &20,000   & 2000        & 2000 \\
 Scenarios & & 670      & 69          & 69 & & 30      & 5          & 5 \\

\bottomrule
\end{tabular}
\caption{Data split statistics for BSD and AMI dataset}
\label{tab:split}
\end{table}

\section {Methodology}
In this section, we analyse our context-aware NMT approach in a dialogue setup in two steps: firstly, we consider what type of information might be useful as context and how it should be encoded to generate useful input representations, and secondly, we discuss modifications in the original encoder-decoder architecture that facilitate learning to attend to context even when tuning on small datasets.

\subsection{Encoding Context}
\label{sec:concat}
\begin{figure}[t!]
    \centering
    \includegraphics[width=\textwidth]{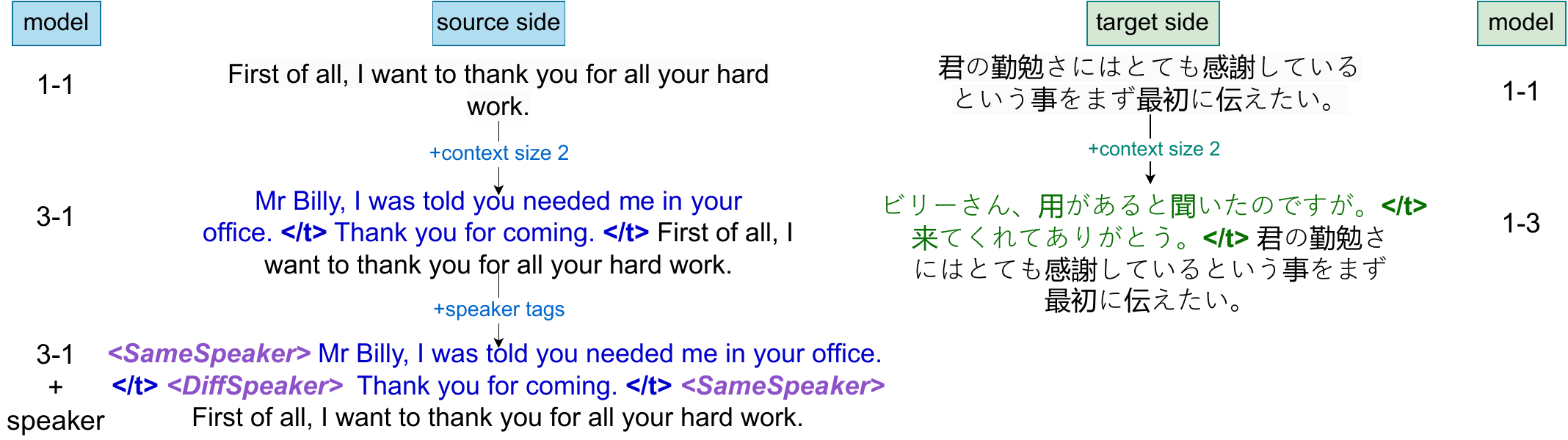}
    \caption{Context-extended inputs on source and target side. Coloured text corresponds to added context, \textbf{bold} signifies context separators and \textbf{\textit{bold-italics}} speaker-related context tags.}
    \label{fig:context}
\end{figure}

We adapt the method of \citet{tiedemann-scherrer-2017-neural} and experiment with encoding contexts both on source-side and target-side.  
Unlike \citet{tiedemann-scherrer-2017-neural} which considers a single preceding sentence, we experiment with up to five preceding sentences, motivated by the findings of \citet{fernandes-etal-2021-measuring,castilho2020context}. We intercept a separator token \texttt{</t>} following every context sentence as shown in Figure \ref{fig:context}.

We compare the context-aware models to the context-agnostic model, finetuned on our dataset.
Henceforth, in this work, we will refer to the context-agnostic model as a 1-1 model, meaning that the model's source-side input is only 1 source sentence, and the target-side input is also only 1 target sentence during the training. 
For the context-aware models, this paper uses the naming convention of 2-1, 3-1, 4-1, and 5-1 for source context-aware models and 1-2, 1-3, 1-4, and 1-5 for target context-aware models. 

Note that in this work we use the gold data (human-generated translations of previous sentences) to represent the target context.
Although the accessibility of target-side context data is limited in real-world translation tasks, there are some relevant use cases.
For example, in a chatbot system where a human can edit the predicted translation in preceding sentences before the current sentence translation, the gold label of preceding target-side sentences is accessible.

{\textbf{Speaker Information: }}Delving deeper into the dialogue scenario, we also  explore whether speaker-related information can provide useful context. 
In a dialogue dataset with multiple speakers, each speaker may utter a varying number of sentences per turn, {and as such using a fixed context window implies potentially including multiple speakers in the context}. 
Since aspects such as discourse style, politeness, {honorifics in Japanese \citep{feely-etal-2019-controlling}} or even topic distribution can be tied to specific speakers, knowing when a speaker changes in the context can be particularly informative. Speaker information has been used to improve user experience in simultaneous interpretation \citep{wang-etal-2022-multimodal-simultaneous}, but to the best of our knowledge, it has not been explored as a contextual feature for MT.

{Hence}, we consider two speaker types: (1) the one who utters the sentence to be translated -- and who may have communicated more sentences in the context window -- (same speaker) and {(2)} any other speaker(s) with utterances within the context window (different speaker), between which  we do not differentiate. 
In other words, we only encode information about whether there has been \textbf{a change of speakers} 
within the context. We achieve this by concatenating 
 either a special token \texttt{<DiffSpeak>} (Different speaker) or a \texttt{<SameSpeak>} (Same speaker)  to each sentence (utterance) of the context as shown in the last row of Figure \ref{fig:context}. This example also highlights the potential difference in speaker formality: the boss uses more casual expressions compared to the employee.

{\textbf{Scene Information: }}
{Similar to} speaker information, {we consider the information associated with the dialogue scene and its potential impact on the translation if used as context. We hence experiment with an additional special token representing the scene tag in BSD dataset}. 
Following BSD dataset scene tags explained in \S\ref{sec:data}, we prepared six additional tokens; \texttt{<face-to-face conversation>}, \texttt{<phone call>}, \texttt{<general chatting>}, \texttt{<meeting>}, \texttt{<training>}, and \texttt{<presentation>}.
One of the tags is concatenated at the very beginning of each source input {to signify} the scene {of the dialogue}. 
For example, the scene tag of conversation in Figure~\ref{fig:context} is \texttt{<face-to-face conversation>}, so 
the 2-1 model's input will be ``\texttt{<face-to-face conversation>} \emph{Thank you for coming. \texttt{</t>} First of all, I want to thank you for all your hard work.''}. 
Such information could provide a useful signal regarding the speaker style, such as honorifics and formality, or even scene-specific terminology. 

\subsection{Context-aware Model Architecture}
\label{sec:model_arc}

To encode context we rely on the \citet{tiedemann-scherrer-2017-neural} approach, which we adapt to optimise performance for the BSD dataset. Due to the small size of available datasets for the business dialogue scenarios it is difficult to train a context-aware transformer architecture from scratch. Instead, we opt for fine-tuning a multi-lingual large pretrained model. 

{\textbf{Baseline: }}All the models for En-Ja translation in this experiment are finetuned with mBART50  \citep{liu-etal-2020-multilingual-denoising, tang2021multilingual} with our proposed architectural modification for context-aware models described in the following paragraphs. We train all models until convergence on the validation set and use a \texttt{max\_token\_length} of size $128$ for the baseline model, and $256$ for the context-aware ones \footnote{All hyperparameters are at: 
\url{https://github.com/su0315/discourse_context_mt}}.
mBART is one of the state-of-the-art multilingual NMT models, with a Transformer-based architecture \citep{NIPS2017_3f5ee243}.
It follows BART \citep{lewis-etal-2020-bart}
Seq2Seq pretraining scheme and is pretrained in 50 languages, including Japanese and English, using multilingual denoising auto-encoder strategy.

{\textbf{Target context-aware model: }}
{To consider context on the target side we essentially decode the target-context as shown in Figure \ref{fig:context} instead of a single sentence.} To apply the \citet{tiedemann-scherrer-2017-neural}'s context-aware approach to the target-side, the baseline model architecture was modified to prevent the {loss function from accounting for mispredicted context and optimising instead only for the original target sentence.} 

\begin{figure}[hb!]
  \begin{minipage}{\textwidth}
  \begin{minipage}[b]{0.49\textwidth}
    \centering     
        \begin{tabular}{ccc}
        \toprule
        Context Size & Baseline & CoAttMask  \\ \midrule
        0 & 0.724  & -  \\ 
        1 & 0.661  & 0.724  \\ 
        2 & 0.665  & 0.724   \\ 
        3 & 0.662 & \textbf{0.727}  \\ 
        4 & 0.658 & \textbf{0.727} \\ \bottomrule
    \end{tabular}
    \captionof{table}{Performance of CoAttMask model in COMET. \textbf{Bold} scores signify the performance improved 1-1 model}
    \label{coatt_score}
    \end{minipage}
  \hfill
    \begin{minipage}[b]{0.49\textwidth}
    \centering
    \includegraphics[width=\textwidth]{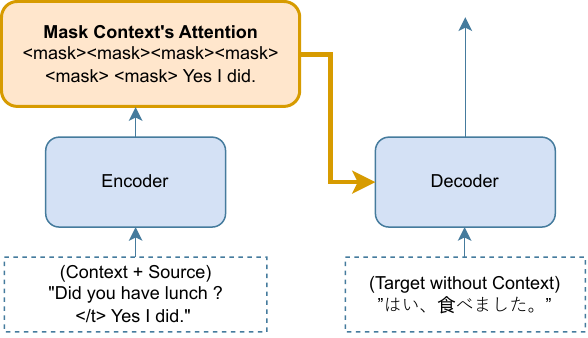}
    \captionof{figure}{CoAttMask Architecture 
    }
    \label{fig:coatt_arc}
  \end{minipage}
  \end{minipage}
\end{figure}

{\textbf{Source context-aware model: }}
{Contrary to \citep{tiedemann-scherrer-2017-neural,bawden-etal-2018-evaluating} we found that directly}  using 
the extended source inputs resulted in significantly lower performance  
for all context sizes, when compared to the original context-agnostic model 
(see Table \ref{coatt_score}). 
{We attribute this inconsistency in our findings to the small size of the BSD dataset which might be insufficient for tuning a large pretrained model towards a context-aware setup.}  

To {address this issue,} 
a new architecture Source \textbf{Co}ntext \textbf{Att}ention \textbf{Mask} Model (CoAttMask) is proposed. In this approach, we pass the context-extended input to the encoder part of the model but mask the encoder outputs that correspond to the context when passed to the decoder.
As shown in the yellow block in Figure \ref{fig:coatt_arc}, after the context-extended input is passed to the encoder, we mask the context-related part when passing the encoded input to the decoder to compute cross attention. As such, the context is leveraged to compute better input representations through self-attention in the transformer but does not further complicate the decoding process. 
Table~\ref{coatt_score} shows that the CoAttMask model successfully outperformed the baseline model architecture (without CoAttMask). 

\section {Evaluation}
\label{sec:evaluation}
\subsection{Metrics for Overall Performance}
\label{sec:intro_eval}
To report the performance of the MT models, we report BLEU \citep{papineni-etal-2002-bleu} and COMET \citep{rei-etal-2020-comet} scores. We use COMET as the primary metric since it has shown to be more efficient in assessing MT quality, better capturing valid synonyms and paraphrases \citep{smith-etal-2016-climbing} as well as discourse phenomena in longer text \citep{10.1145/3441691}. 

\subsection{Metric for Context Usage {-- CXMI --}}
Although COMET can capture more semantic features than BLEU, it is still difficult to assess how much context-aware NMT models actually use the additional contexts to improve predictions. To that end, we use Conditional Cross Mutual Information (CXMI) \citep{bugliarello-etal-2020-easier,fernandes-etal-2021-measuring}. CXMI measures the entropy (information gain) of a context-agnostic machine translation model and a context-aware machine translation model.
The CXMI formula can be seen in Eq. (1), where ${C}$ signifies additional context,  ${Y}$ the target, ${X}$ the source, ${H_{qMT_{A}}}$ the entropy of a context-agnostic
machine translation model, and ${H_{qMT_{C}}}$ the entropy of context-aware machine translation model.
Thus, a positive CXMI score indicates a useful contribution of context to predicting the correct target (increasing the predicted score of the correct target words).
This can be estimated with Eq. (2), over a test dataset with $N$ sentences, when $y^{(i)}$ is $i$\textsuperscript{th} target sentence and $x^{(i)}$ the $i$\textsuperscript{th} source sentence in each document \citep{fernandes-etal-2021-measuring}. 
\begin{align}
CXMI \left (C \to  Y|X\right) 
&= H_{q_{MT_{A}}}\left ( Y|X \right)- H_{q_{MT_{C}}}\left ( Y|X, C \right) \\
&\approx  -  \frac{1}{N}  \sum_{i=1}^N \log \frac{q_{MT_{A}}( y^{(i)}|x^{(i)} ) }{{q_{MT_{C}}( y^{(i)}|x^{(i)}, C^{(i)} ) }} 
\end{align}

In this experiment, CXMI is calculated between context-aware models with preceding sentence(s), speaker information, and scene information and each corresponding baseline model that lacks the respective context.   
To compute CXMI, a single model that can be tested with both context-agnostic inputs and context-extended inputs is required. {We hence train the models with dynamic context size, such that during training the model can see anywhere from 0 to $k$ context sentences} \citep{fernandes-etal-2021-measuring}.

\subsection{\emph{Honorifics P-CXMI}}
\label{sec:hon}

To evaluate how much additional context is actually used to improve translation with respect to honorifics, we also  compute P-CXMI, an extension of CXMI that allows us to measure the impact of context on specific translations or words in a translation instead of over the whole corpus \citep{yin2021does}. We define \emph{Honorifics P-CXMI} for token-level honorific expressions, which we calculate 
only for cases where the gold label is an honorific expression. 
While CXMI is calculated on the corpus level, 
 averaged over the number of sentences, \emph{Honorifics P-CXMI}  is calculated for each honorific token and averaged over the number of the honorific tokens in the testset. As such, it is not directly comparable to the CXMI values~\citep{yin2021does}.

Inspired by Japanese honorific word lists proposed in \citet{yin2021does} 
and \citet{farajian-etal-2020-findings}, the following tokens are selected as the main honorific expressions {(based on frequency of use and non-ambiguous functionality in the sentence)} \footnote{Modified for the mBART50 tokenizer.} 
``です (desu)'', ``でした (deshita)'', ``ます (masu)'', ``ました (mashita)'', ``ません (masen)'', ``ましょう (mashou)'',``でしょう 
(deshou)'',``ください (kudasai)'',``ございます (gozaimasu)'',``おります(orimasu)'', ``致します (itashimasu)'', ``ご覧 (goran)'', ``なります (narimasu)'', ``伺 (ukaga)'', ``頂く (itadaku)'', ``頂き (itadaki)'', ``頂いて (itadaite)'', ``下さい (kudasai)'', ``申し上げます (moushiagemasu)''.
Those tokens are mainly categorized as three types of honorifics: respectful (sonkeigo, 尊敬語), humble (kenjogo, 謙譲語), polite (teineigo, 丁寧語).

\section{Experimental Results}
We compare our work to previous approaches evaluated on BSD, namely this of 
\cite{rikters-etal-2019-designing} who combined multiple En-Ja datasets to train a model for En-Ja dialogue translation and \cite{rikters2021japanese} who also used a context-aware variant of \cite{tiedemann-scherrer-2017-neural} combined with factors to encode dialogue context. Additionally, we compare with our context agnostic baseline. 
{Table~\ref{src_tgt_score} shows that {tuning mBART on the BSD data already} outperformed the previous studies by more than 9 points in terms of  BLEU, highlighting the impact of pretraining on large multilingual data}.
For the context-aware models, four types of  models are compared for different context sizes; (1) Preceding Sentences Model (\S\ref{sec:preceding_result}); (2) Speaker Information Model; 3) Scene Information Model; and (4) Speaker \& Scene Information Model (\S\ref{sec:extra_result}).%

\subsection{Context-aware Models: Preceding Sentences}
\label{sec:preceding_result}
As seen in Table~\ref{src_tgt_score}, as we increase the size of the context used, the CXMI score consistently increases indicating better leveraging of the context provided for the prediction of the target words. However, this increased attention to context is only reflected in small gains in the overall performance for specific context sizes. Specifically, for the source-side context only the models with larger context of 3 and 4 sentences improved for BLEU and COMET, as opposed to previous work that observes gains on single sentence context 
and often decreasing performance for larger context sizes 
\citep{tiedemann-scherrer-2017-neural, voita-etal-2018-context, rikters-EtAl:2020:WMT, ri-etal-2021-zero, nagata-morishita-2020-test}. We hypothesize that this relates to our stronger baseline, and the specifics of the dialogue translation task: shorter utterances on average and multiple speakers which could lead to useful context lying further away in the dialogue history. 

\begin{figure}[t!]
\begin{minipage}{\columnwidth}
  \begin{minipage}[b]{0.49\columnwidth}
    \centering
    \includegraphics[width=\columnwidth]{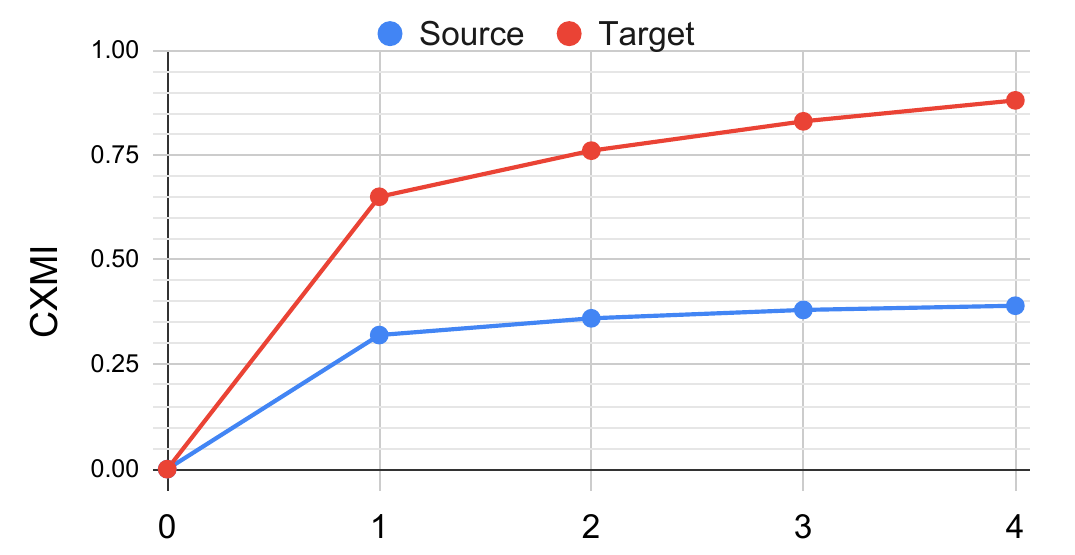}
    \captionof{figure}{CXMI for source and target context-aware models in each context size}
    \label{fig:src_tgt_cxmi}
  \end{minipage}
  \hfill
  \begin{minipage}[b]{0.49\columnwidth}
    \centering     
        \includegraphics[width=\columnwidth]{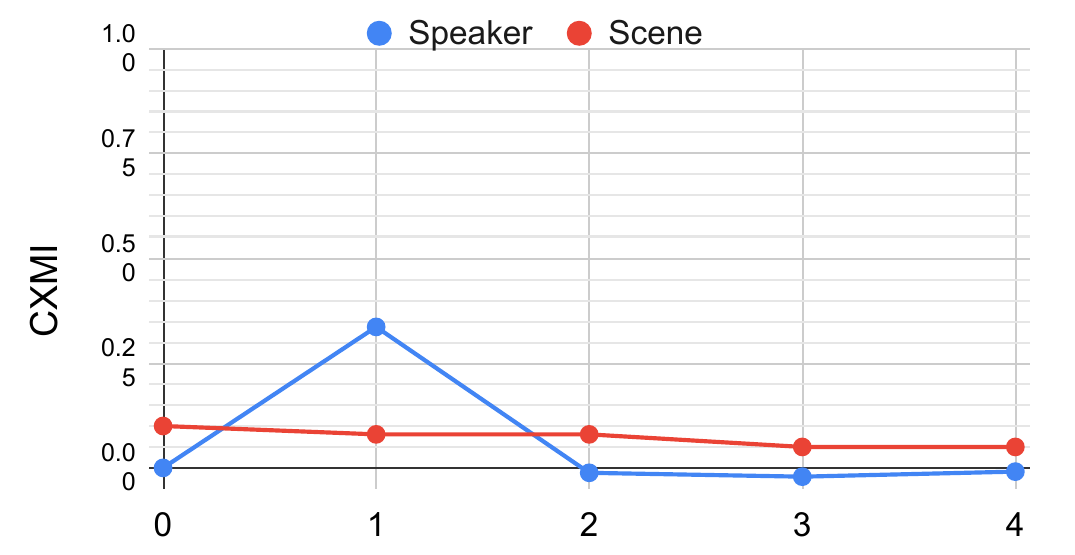}
        \captionof{figure}{CXMI for speaker and scene model in each context size}
        \label{fig:speaker_scene_cxmi}
    \end{minipage}
  \end{minipage}
\end{figure}

For the target-side context most variants either under-performed or performed similarly to the context-agnostic model. 
Indeed, while we notice an increased usage of context as we increase the target context size (see Figure \ref{fig:src_tgt_cxmi}), this does not seem to lead to improved performance. Further supported by the findings in \S\ref{sec:ami} on the AMI dataset, it seems that using context on the source side is more beneficial for such small dialogue datasets and we focus our analysis and experiments more on the source side. However, it would be interesting to consider further adapting target-side context or explore pre-training on larger corpora as a way to mitigate this in future work~\citep{liang2022scheduled,Su2019Exploiting}.
\begin{table}[t!]
    \centering
    \resizebox{0.7\columnwidth}{!}{%
    \begin{tabular}{lllll}
\toprule
& \begin{tabular}[c]{@{}l@{}}Model (context size)\end{tabular} & \multicolumn{1}{l}{BLEU $\uparrow$} & \multicolumn{1}{l}{COMET $\uparrow$} & \multicolumn{1}{l}{CXMI $\uparrow$} \\ \midrule
&\citet{rikters-etal-2019-designing} (0)& 13.53 & - & -\\ 
&        \citet{rikters2021japanese} (0)& 12.93 & - & - \\ 
&        \citet{rikters2021japanese} (1)& 14.52 & - & - \\ 
& \citet{ri-etal-2021-zero} (1)& 17.11 & - & - \\
\multirow{-5}{*}{Baselines}&1-1 (0)     & 26.04 & 0.725  & 0 \\ \midrule 
        &2-1 (1) & 25.87 & 0.724 & {0.32} \\ 
        &3-1 (2) & 25.41 & 0.724 & {0.36}  \\ 
        &4-1 (3) & \textbf{26.09} & \textbf{0.727} & {0.38}  \\ 
        \multirow{-4}{*}{\begin{tabular}[c]{@{}l@{}}Source \\ context \end{tabular}}        
        &5-1 (4) & \textbf{26.09} & \textbf{0.727} & {0.39}\\  \midrule
        &1-2 (1) & 25.85 & 0.72 & {0.65}  \\ 
        &1-3 (2) & \textbf{26.08} & 0.702  & {0.76}  \\ 
        &1-4 (3) & 25.77 & 0.704 & {0.83}\\ 
    \multirow{-4}{*}{\begin{tabular}[c]{@{}l@{}}Target \\ context \end{tabular}}  
        &1-5 (4) & 24.96 & 0.71 & {0.88} \\ \bottomrule
    \end{tabular}%
    }
    \caption{Score comparison between preceding sentences models and 1-1 model. \textbf{Bold} scores signify the performance improved baseline (BLEU, COMET) }
    \label{src_tgt_score}
\end{table}

Focusing on CXMI as shown in Table~\ref{src_tgt_score} and Figure~\ref{fig:src_tgt_cxmi}, our experiments corroborate the main findings of \cite{fernandes-etal-2021-measuring}. We can see that for both target and source the biggest jump in context usage is when we increase the context size from 0
to 1, but unlike \cite{fernandes-etal-2021-measuring} we subsequently observe small but consistent increases for each context size (ascending).

Table~\ref{tab:hon_preceding} 
shows the result of \emph{Honorifics CXMI} between source-side preceding sentences models and 1-1 model.
With respect to the translation of honorifics, 
\emph{Honorifics CXMI} scores for all context sizes show positive score, indicating that the provision of additional context helps the model to attribute higher density to the correct honorific translation. In other words, the model can leverage additional context to improve the prediction of honorific expressions.

Looking at the improved scores for each context size and honorific expression separately, we found that in all cases, it was the translation of the honorific token ``伺 (ukaga)'' that benefited the most. 
``伺 (ukaga)'' is an honorific token that is a component of ``伺う(ukagau)'', a verb meaning ``go'' or ``ask'' in Japanese honorific expression.
In particular, ``伺う(ukagau)'' is one of the humble (kenjogo, 謙譲語) expressions, and the humble is 
used in a business email or very formal speech \citep{liu-kobayashi-2022-construction}. 
These honorific expressions are used strictly by speakers to refer to themselves when they address a superior in business settings \citep{rahayu2013japanese}. As such, previous utterances that would reveal the relation of the speaker to the addressee are necessary to obtain the correct translation.
Table~\ref{tab:ukagau} demonstrates the correction in the use of  ``伺 (ukaga)'' when using a context window of size 2. 
The baseline model predicts ``申します'' instead of ``伺 (ukaga)'', leading to a semantically inappropriate translation meaning ``\emph{I'm (Takada)}'' while with additional context it correctly predicts the ``伺 (ukaga)'' token. 

\begin{table}[t!]
\centering
\begin{tabular}{ccccc}
\toprule
&2-1&3-1&4-1&5-1 \\ \midrule
 {{\begin{tabular}[c]{@{}l@{}}\emph{Honorifics} \emph{CXMI} $\uparrow$\end{tabular}}} 
&0.05 & 0.07 & 0.06 & 0.06 
\\ \bottomrule
\end{tabular}
\caption{\emph{Honorifics CXMI} between source-side preceding sentences models and 1-1 model}
\label{tab:hon_preceding}
\end{table}
\begin{table*}[t!]
	\resizebox{1\linewidth}{!}{%
		\begin{tabular}{p{3cm} p{3cm} p{3cm} p{3cm}}
			\toprule
			Source Sentence  & Reference Sentence & 1-1 Model Prediction  & 3-1 Model Prediction \\ \midrule
			\small{I, Takada from Company I will \underline{go} to your place
				at 5 o'clock in the afternoon tomorrow.} & \small{明日の午後５時に、わたくし、I社の高田が\underline{伺います}。} & \small{明日の午後 5時に、I社の高田 と申します。}  & \small{私、I社の高田が明 日の午後5時に御 社へお\underline{伺いします}。}   \\ \bottomrule
		\end{tabular}
	}
	\caption{Comparison between a context-agnostic model (1-1) and a context-aware model (3-1) in predicting honorific token ``伺''.  (\underline{Underlined} words signify that the 3-1 model improved the 1-1 model in predicting the correct token.)}
	\label{tab:ukagau}
\end{table*}

\subsection{Extra-sentential context:}
\label{sec:extra_result}
For the following experiments, we focus on further enhancing the source-side context by adding scene and speaker information as discussed in \S\ref{sec:concat}. We first explore their usefulness separately, concatenating to the context either speaker tags or scene tags, as shown in Table~\ref{tab:speaker_score} and Figure \ref{fig:speaker_scene_cxmi}.

{\textbf{Speaker Information Models: }}
When adding speaker information (``With Speaker'', Table \ref{tab:speaker_score}) the model seems to be obtaining slightly better performance on BLEU scores but not COMET. Additionally, with respect to the CXMI (see Figure \ref{fig:speaker_scene_cxmi}), the speaker information seems to be useful for the model predictions only when using a single sentence of context. In other words, the model benefits only from knowing whether the previous utterance originated from the same speaker or not. While this finding is quite intuitive (a change of speaker could indicate a switch in style and formality) it is still unclear why this does not hold for larger context windows. 

Note that while the benefits of using the speaker turn information seem limited, there are further aspects to be explored that were out of scope in this work. Specifically, given sufficient training data one could use a separate tag for each speaker in case of $\leq 2$ speakers, either using abstract speaker tags, or even the speaker names, potentially helping toward pronoun translation.

\begin{table}[htb!]
\centering
\resizebox{\linewidth}{!}{%
\begin{tabular}{ccccccccc}
\toprule
& \multicolumn{2}{c}{Preceding Sentences} & \multicolumn{2}{c}{With Speaker}   & \multicolumn{2}{c}{With Scene}  & \multicolumn{2}{c}{With Speaker \& Scene}  \\ \cmidrule(l{2pt}r{2pt}){2-3} \cmidrule(l{2pt}r{2pt}){4-5} \cmidrule(l{2pt}r{2pt}){6-7}  \cmidrule(l{2pt}r{2pt}){8-9}
Model (Context Size) & {BLEU$\uparrow$}    & COMET$\uparrow$  & {BLEU$\uparrow$}    & {COMET$\uparrow$}   & {BLEU$\uparrow$}  & {COMET$\uparrow$} & {BLEU$\uparrow$}  & {COMET$\uparrow$}  \\ \midrule
1-1 (0)          & 26.04       &   0.725     & -        & -            & \textbf{26.19} & \textbf{0.726} & - & - \\ 
2-1 (1)         & 25.87 & 0.724 & {{25.94}} & {0.718}    & {\textbf{26.18}} & {{0.727}} & \textbf{26.18} & \textbf{0.730}\\ 
3-1 (2)         & {25.41}   & 0.724 & {{26.09}} & {0.722}  & {{26.26}} & {{0.727}} & \underline{\textbf{26.41}} & \underline{\textbf{0.740}} \\ 
4-1 (3)         & {26.09}  & 0.727  & {26.03} & {0.722}  & {\textbf{26.27}} & {\textbf{0.731}} & 26.07 & 0.730 \\ 
5-1 (4)         & {26.09} & 0.727 & {\textbf{26.39}} & {0.726} & {{26.1}}  & {\textbf{0.728}} & 26.15 & 0.720 \\ \bottomrule
\end{tabular}%
}
\caption{Score comparison among preceding sentence models (w/o speaker and scene information), and models with addition of speaker and scene tags. \textbf{Bold} scores signify the best performance for each context size and \underline{underlined} ones the best performance overall.}
\label{tab:speaker_score}
\end{table}

{\bf{Scene Information Model: }}
Unlike the speaker information, scene information can be added when the context size is zero too, since it does not need preceding sentences. 

In contrast to speaker information models, ``With Scene'' models outperformed ``Preceding Sentences'' models for both BLEU and COMET on all context sizes, including when used with no additional context.
Additionally, CXMI remains positive for all context sizes with a small decrease when the context size is larger. Hence, we can conclude that scene information helps towards the correct translation especially when limited context is available.

{\textbf{Speaker and Scene Model: }}
We finally investigate if combining scene and speaker information can further improve performance.
Indeed, for smaller context windows (speaker \& scene models  2-1 and 3-1) outperformed their respective scene-only and speaker-only versions.  
Also, the 3-1 speaker \& scene model obtained the best performance overall. 
Hence, while speaker information on its own did not improve performance, the combination of speaker information and scene information outperformed the models without them. 
This finding indicates that for specific scenarios (scenes), speaker turn might provide more useful signal. Indeed, depending on the scene the speakers may change more or less frequently signifying a necessary change of style (e.g. compare a presentation scene versus the phone call one). 
It would be interesting to further explore the relationship between the speaker switch frequency and scene type in the future.

\subsection{Performance on the AMI dataset}
\label{sec:ami}
To examine the context-aware model's performance on a similar dataset, we also tested the trained preceding sentences models using AMI dataset introduced in \S\ref{sec:data}.
Table~\ref{tab:ami-score} shows the performance of the context-aware models on increasing context size.
Both context-aware and context-agnostic models obtain higher scores on the AMI dataset, compared to BSD. We notice however that we obtain small performance boosts for some context-aware combinations. More importantly, CXMI findings corroborate those on BSD: as the context size gets larger, CXMI increases both on source and target side. The similar CXMI trends reinforce our findings, hinting that they are not artifacts of a specific dataset, but rather a property of the language pair.

\begin{table}[ht!]
 \centering
 \resizebox{0.9\textwidth}{!}{%
\begin{tabular}{cccccccccc}
\toprule
& Baseline & \multicolumn{4}{c}{Source Side}& \multicolumn{4}{c}{Target Side} \\
    \cmidrule(l{2pt}r{2pt}){3-6} \cmidrule(l{2pt}r{2pt}){7-10}\cmidrule(l{2pt}r{2pt}){2-2} 
      & 1-1   & 2-1   & 3-1   & 4-1    & 5-1   & 1-2   & 1-3   & 1-4   & 1-5   \\
      \midrule
BLEU  & 32.46 & \textbf{32.8}  & 32.12 & \textbf{32.61}  & 32.05 & 32.13 & 31.22 & 31.29 & \textbf{32.56} \\
COMET & 0.852 & \textbf{0.858} & 0.846 & \textbf{0.854} & 0.846 & 0.848 & 0.833 & 0.833 & 0.85  \\ 
CXMI  & - & 0.24  & 0.27  & 0.31   & 0.34  & 0.07  & 0.17  & 0.25  & 0.48  \\ \bottomrule
\end{tabular}%
}
\caption{Score comparison between preceding sentences models and 1-1 models with AMI dataset. \textbf{Bold} scores signify the performance improved over the baseline (BLEU, COMET).}
\label{tab:ami-score}
\end{table}

\section{Conclusion and Future Work}
This paper explored to what degree encoded context can improve NMT performance for English-Japanese dialogue translation, and what kind of context provides useful information.
With our proposed method,
we were able to tune mBART on small dialogue datasets and obtain improved MT performance using context. We found that source-side context was more beneficial towards performance and that complementing our source-side context with scene and speaker-turn tags provided further performance improvements. We further analyse the impact of our proposed context-aware methods on the translations obtained, with a focus on translation of Japanese honorifics.
In future work, we aim to further investigate context for dialogue translation, expanding to a multilingual setup, larger datasets, and additional extra-sentential context. 

\section*{Acknowledgements}
This work was supported by EU's
Horizon Europe Research and Innovation Actions
(UTTER, contract 101070631), by
the Portuguese Recovery and Resilience Plan
through project C645008882-00000055 (NextGenAI, Center for Responsible AI), and by Computational Linguistics, University of Potsdam, Germany. 

\small

\bibliographystyle{apalike}
\bibliography{mtsummit2023}

\begin{thebibliography}{}

\bibitem[Agrawal et~al., 2018]{agrawal2018contextual}
Agrawal, R.~R., Turchi, M., and Negri, M. (2018).
\newblock Contextual handling in neural machine translation: Look behind, ahead
  and on both sides.
\newblock In {\em Proceedings of the 21st Annual Conference of the European
  Association for Machine Translation}, pages 11--20.

\bibitem[Bawden et~al., 2018]{bawden-etal-2018-evaluating}
Bawden, R., Sennrich, R., Birch, A., and Haddow, B. (2018).
\newblock Evaluating discourse phenomena in neural machine translation.
\newblock In {\em Proceedings of the 2018 Conference of the North {A}merican
  Chapter of the Association for Computational Linguistics: Human Language
  Technologies, Volume 1 (Long Papers)}, pages 1304--1313, New Orleans,
  Louisiana. Association for Computational Linguistics.

\bibitem[Bugliarello et~al., 2020]{bugliarello-etal-2020-easier}
Bugliarello, E., Mielke, S.~J., Anastasopoulos, A., Cotterell, R., and Okazaki,
  N. (2020).
\newblock It{'}s easier to translate out of {E}nglish than into it: {M}easuring
  neural translation difficulty by cross-mutual information.
\newblock In {\em Proceedings of the 58th Annual Meeting of the Association for
  Computational Linguistics}, pages 1640--1649, Online. Association for
  Computational Linguistics.

\bibitem[Castilho et~al., 2020]{castilho2020context}
Castilho, S., Popovi{\'c}, M., and Way, A. (2020).
\newblock On context span needed for machine translation evaluation.
\newblock In {\em Proceedings of the Twelfth Language Resources and Evaluation
  Conference}, pages 3735--3742, Marseille, France. European Language Resources
  Association.

\bibitem[Chen et~al., 2021]{chen2021improving}
Chen, L., Li, J., Gong, Z., Duan, X., Chen, B., Luo, W., Zhang, M., and Zhou,
  G. (2021).
\newblock Improving context-aware neural machine translation with source-side
  monolingual documents.
\newblock In {\em IJCAI}, pages 3794--3800.

\bibitem[Farajian et~al., 2020]{farajian-etal-2020-findings}
Farajian, M.~A., Lopes, A.~V., Martins, A. F.~T., Maruf, S., and Haffari, G.
  (2020).
\newblock Findings of the {WMT} 2020 shared task on chat translation.
\newblock In {\em Proceedings of the Fifth Conference on Machine Translation},
  pages 65--75, Online. Association for Computational Linguistics.

\bibitem[Feely et~al., 2019]{feely-etal-2019-controlling}
Feely, W., Hasler, E., and de~Gispert, A. (2019).
\newblock Controlling {J}apanese honorifics in {E}nglish-to-{J}apanese neural
  machine translation.
\newblock In {\em Proceedings of the 6th Workshop on Asian Translation}, pages
  45--53, Hong Kong, China. Association for Computational Linguistics.

\bibitem[Fernandes et~al., 2023]{yin2021does}
Fernandes, P., Yin, K., Martins, A.~F., and Neubig, G. (2023).
\newblock When does translation require context? a data-driven, multilingual
  exploration.
\newblock {\em Proceedings of the 61st Annual Meeting of the Association for
  Computational Linguistics}.

\bibitem[Fernandes et~al., 2021]{fernandes-etal-2021-measuring}
Fernandes, P., Yin, K., Neubig, G., and Martins, A. F.~T. (2021).
\newblock Measuring and increasing context usage in context-aware machine
  translation.
\newblock In {\em Proceedings of the 59th Annual Meeting of the Association for
  Computational Linguistics and the 11th International Joint Conference on
  Natural Language Processing (Volume 1: Long Papers)}, pages 6467--6478,
  Online. Association for Computational Linguistics.

\bibitem[Fukada and Asato, 2004]{fukada2004universal}
Fukada, A. and Asato, N. (2004).
\newblock Universal politeness theory: application to the use of japanese
  honorifics.
\newblock {\em Journal of pragmatics}, 36(11):1991--2002.

\bibitem[Hwang et~al., 2021]{hwang2021context}
Hwang, Y., Kim, Y., and Jung, K. (2021).
\newblock Context-aware neural machine translation for korean honorific
  expressions.
\newblock {\em Electronics}, 10(13):1589.

\bibitem[Lewis et~al., 2020]{lewis-etal-2020-bart}
Lewis, M., Liu, Y., Goyal, N., Ghazvininejad, M., Mohamed, A., Levy, O.,
  Stoyanov, V., and Zettlemoyer, L. (2020).
\newblock {BART}: Denoising sequence-to-sequence pre-training for natural
  language generation, translation, and comprehension.
\newblock In {\em Proceedings of the 58th Annual Meeting of the Association for
  Computational Linguistics}, pages 7871--7880, Online. Association for
  Computational Linguistics.

\bibitem[Liang et~al., 2021a]{liang2021modeling}
Liang, Y., Meng, F., Chen, Y., Xu, J., and Zhou, J. (2021a).
\newblock Modeling bilingual conversational characteristics for neural chat
  translation.
\newblock In {\em Proceedings of the 59th Annual Meeting of the Association for
  Computational Linguistics and the 11th International Joint Conference on
  Natural Language Processing (Volume 1: Long Papers)}, pages 5711--5724.

\bibitem[Liang et~al., 2022]{liang2022scheduled}
Liang, Y., Meng, F., Xu, J., Chen, Y., and Zhou, J. (2022).
\newblock Scheduled multi-task learning for neural chat translation.
\newblock In {\em Proceedings of the 60th Annual Meeting of the Association for
  Computational Linguistics (Volume 1: Long Papers)}, pages 4375--4388.

\bibitem[Liang et~al., 2021b]{liang2021towards}
Liang, Y., Zhou, C., Meng, F., Xu, J., Chen, Y., Su, J., and Zhou, J. (2021b).
\newblock Towards making the most of dialogue characteristics for neural chat
  translation.
\newblock In {\em Proceedings of the 2021 Conference on Empirical Methods in
  Natural Language Processing}, pages 67--79.

\bibitem[Liu and Kobayashi, 2022]{liu-kobayashi-2022-construction}
Liu, M. and Kobayashi, I. (2022).
\newblock Construction and validation of a {J}apanese honorific corpus based on
  systemic functional linguistics.
\newblock In {\em Proceedings of the Workshop on Dataset Creation for
  Lower-Resourced Languages within the 13th Language Resources and Evaluation
  Conference}, pages 19--26, Marseille, France. European Language Resources
  Association.

\bibitem[Liu et~al., 2021]{2021}
Liu, S., Sun, Y., and Wang, L. (2021).
\newblock Recent advances in dialogue machine translation.
\newblock {\em Information}, 12(11):484.

\bibitem[Liu et~al., 2020]{liu-etal-2020-multilingual-denoising}
Liu, Y., Gu, J., Goyal, N., Li, X., Edunov, S., Ghazvininejad, M., Lewis, M.,
  and Zettlemoyer, L. (2020).
\newblock Multilingual denoising pre-training for neural machine translation.
\newblock {\em Transactions of the Association for Computational Linguistics},
  8:726--742.

\bibitem[Maruf et~al., 2019]{maruf2019selective}
Maruf, S., Martins, A.~F., and Haffari, G. (2019).
\newblock Selective attention for context-aware neural machine translation.
\newblock In {\em Proceedings of the 2019 Conference of the North American
  Chapter of the Association for Computational Linguistics: Human Language
  Technologies, Volume 1 (Long and Short Papers)}, pages 3092--3102.

\bibitem[Maruf et~al., 2021]{10.1145/3441691}
Maruf, S., Saleh, F., and Haffari, G. (2021).
\newblock A survey on document-level neural machine translation: Methods and
  evaluation.
\newblock {\em ACM Comput. Surv.}, 54(2).

\bibitem[Mino et~al., 2020]{mino-etal-2020-effective}
Mino, H., Ito, H., Goto, I., Yamada, I., and Tokunaga, T. (2020).
\newblock Effective use of target-side context for neural machine translation.
\newblock In {\em Proceedings of the 28th International Conference on
  Computational Linguistics}, pages 4483--4494, Barcelona, Spain (Online).
  International Committee on Computational Linguistics.

\bibitem[Nagata and Morishita, 2020]{nagata-morishita-2020-test}
Nagata, M. and Morishita, M. (2020).
\newblock A test set for discourse translation from {J}apanese to {E}nglish.
\newblock In {\em Proceedings of the Twelfth Language Resources and Evaluation
  Conference}, pages 3704--3709, Marseille, France. European Language Resources
  Association.

\bibitem[Papineni et~al., 2002]{papineni-etal-2002-bleu}
Papineni, K., Roukos, S., Ward, T., and Zhu, W.-J. (2002).
\newblock {B}leu: a method for automatic evaluation of machine translation.
\newblock In {\em Proceedings of the 40th Annual Meeting of the Association for
  Computational Linguistics}, pages 311--318, Philadelphia, Pennsylvania, USA.
  Association for Computational Linguistics.

\bibitem[Rahayu, 2013]{rahayu2013japanese}
Rahayu, E.~T. (2013).
\newblock The japanese keigo verbal marker.
\newblock {\em Advances in Language and Literary Studies}, 4(2):104--111.

\bibitem[Rei et~al., 2020]{rei-etal-2020-comet}
Rei, R., Stewart, C., Farinha, A.~C., and Lavie, A. (2020).
\newblock {COMET}: A neural framework for {MT} evaluation.
\newblock In {\em Proceedings of the 2020 Conference on Empirical Methods in
  Natural Language Processing (EMNLP)}, pages 2685--2702, Online. Association
  for Computational Linguistics.

\bibitem[Ri et~al., 2021]{ri-etal-2021-zero}
Ri, R., Nakazawa, T., and Tsuruoka, Y. (2021).
\newblock Zero-pronoun data augmentation for {J}apanese-to-{E}nglish
  translation.
\newblock In {\em Proceedings of the 8th Workshop on Asian Translation
  (WAT2021)}, pages 117--123, Online. Association for Computational
  Linguistics.

\bibitem[Rikters et~al., 2019]{rikters-etal-2019-designing}
Rikters, M., Ri, R., Li, T., and Nakazawa, T. (2019).
\newblock Designing the business conversation corpus.
\newblock In {\em Proceedings of the 6th Workshop on Asian Translation}, pages
  54--61, Hong Kong, China. Association for Computational Linguistics.

\bibitem[Rikters et~al., 2020]{rikters-EtAl:2020:WMT}
Rikters, M., Ri, R., Li, T., and Nakazawa, T. (2020).
\newblock Document-aligned japanese-english conversation parallel corpus.
\newblock In {\em Proceedings of the Fifth Conference on Machine Translation},
  pages 637--643, Online. Association for Computational Linguistics.

\bibitem[Rikters et~al., 2021]{rikters2021japanese}
Rikters, M., Ri, R., Li, T., and Nakazawa, T. (2021).
\newblock Japanese--english conversation parallel corpus for promoting
  context-aware machine translation research.
\newblock {\em Journal of Natural Language Processing}, 28(2):380--403.

\bibitem[Sennrich et~al., 2016]{sennrich2016controlling}
Sennrich, R., Haddow, B., and Birch, A. (2016).
\newblock Controlling politeness in neural machine translation via side
  constraints.
\newblock In {\em Proceedings of the 2016 Conference of the North American
  Chapter of the Association for Computational Linguistics: Human Language
  Technologies}, pages 35--40.

\bibitem[Smith et~al., 2016]{smith-etal-2016-climbing}
Smith, A., Hardmeier, C., and Tiedemann, J. (2016).
\newblock Climbing mont {BLEU}: The strange world of reachable high-{BLEU}
  translations.
\newblock In {\em Proceedings of the 19th Annual Conference of the {E}uropean
  Association for Machine Translation}, pages 269--281.

\bibitem[Su et~al., 2019]{Su2019Exploiting}
Su, J., Zhang, X., Lin, Q., Qin, Y., Yao, J., and Liu, Y. (2019).
\newblock Exploiting reverse target-side contexts for neural machine
  translation via asynchronous bidirectional decoding.
\newblock {\em Artificial Intelligence}, 277:103168.

\bibitem[Sugiyama and Yoshinaga, 2019]{sugiyama-yoshinaga-2019-data}
Sugiyama, A. and Yoshinaga, N. (2019).
\newblock Data augmentation using back-translation for context-aware neural
  machine translation.
\newblock In {\em Proceedings of the Fourth Workshop on Discourse in Machine
  Translation (DiscoMT 2019)}, pages 35--44, Hong Kong, China. Association for
  Computational Linguistics.

\bibitem[Tang et~al., 2021]{tang2021multilingual}
Tang, Y., Tran, C., Li, X., Chen, P.-J., Goyal, N., Chaudhary, V., Gu, J., and
  Fan, A. (2021).
\newblock Multilingual translation from denoising pre-training.
\newblock In {\em Findings of the Association for Computational Linguistics:
  ACL-IJCNLP 2021}, pages 3450--3466.

\bibitem[Tiedemann and Scherrer, 2017]{tiedemann-scherrer-2017-neural}
Tiedemann, J. and Scherrer, Y. (2017).
\newblock Neural machine translation with extended context.
\newblock In {\em Proceedings of the Third Workshop on Discourse in Machine
  Translation}, pages 82--92, Copenhagen, Denmark. Association for
  Computational Linguistics.

\bibitem[Tu et~al., 2018]{tu-etal-2018-learning}
Tu, Z., Liu, Y., Shi, S., and Zhang, T. (2018).
\newblock Learning to remember translation history with a continuous cache.
\newblock {\em Transactions of the Association for Computational Linguistics},
  6:407--420.

\bibitem[Vaswani et~al., 2017]{NIPS2017_3f5ee243}
Vaswani, A., Shazeer, N., Parmar, N., Uszkoreit, J., Jones, L., Gomez, A.~N.,
  Kaiser, L.~u., and Polosukhin, I. (2017).
\newblock Attention is all you need.
\newblock In Guyon, I., Luxburg, U.~V., Bengio, S., Wallach, H., Fergus, R.,
  Vishwanathan, S., and Garnett, R., editors, {\em Advances in Neural
  Information Processing Systems}, volume~30. Curran Associates, Inc.

\bibitem[Voita et~al., 2019]{voita2019good}
Voita, E., Sennrich, R., and Titov, I. (2019).
\newblock When a good translation is wrong in context: Context-aware machine
  translation improves on deixis, ellipsis, and lexical cohesion.
\newblock In {\em Proceedings of the 57th Annual Meeting of the Association for
  Computational Linguistics}, pages 1198--1212.

\bibitem[Voita et~al., 2018]{voita-etal-2018-context}
Voita, E., Serdyukov, P., Sennrich, R., and Titov, I. (2018).
\newblock Context-aware neural machine translation learns anaphora resolution.
\newblock In {\em Proceedings of the 56th Annual Meeting of the Association for
  Computational Linguistics (Volume 1: Long Papers)}, pages 1264--1274,
  Melbourne, Australia. Association for Computational Linguistics.

\bibitem[Wang et~al., 2022]{wang-etal-2022-multimodal-simultaneous}
Wang, X., Utiyama, M., and Sumita, E. (2022).
\newblock A multimodal simultaneous interpretation prototype: Who said what.
\newblock In {\em Proceedings of the 15th Biennial Conference of the
  Association for Machine Translation in the Americas (Volume 2: Users and
  Providers Track and Government Track)}, pages 132--143, Orlando, USA.
  Association for Machine Translation in the Americas.

\bibitem[Xu et~al., 2021]{xu2021efficient}
Xu, H., Xiong, D., Van~Genabith, J., and Liu, Q. (2021).
\newblock Efficient context-aware neural machine translation with layer-wise
  weighting and input-aware gating.
\newblock In {\em Proceedings of the Twenty-Ninth International Conference on
  International Joint Conferences on Artificial Intelligence}, pages
  3933--3940.

\end{thebibliography}

\end{document}